# A Type II Singularity Avoidance Algorithm for Parallel Manipulators using Output Twist Screws

José L. Pulloquinga[a,∗], Rafael J. Escarabajal[a], Ángel Valera[a], Marina Vallés[a], Vicente Mata[b]

[a]*Departamento de Ingeniería de Sistemas y Automatica, Universitat Politècnica de València, Camino de Vera, s/n, Valencia, 46022, Valencia, Spain*
[b]*Departamento de Ingeniería Mecánica y de Materiales, Universitat Politècnica de València, Camino de Vera, s/n, Valencia, 46022, Valencia, Spain*

**Abstract**

Parallel robots (PRs) are closed-chain manipulators with diverse applications due to their accuracy and high payload. However, there are configurations within the workspace named Type II singularities where the PRs lose control of the end-effector movements. Type II singularities are a problem for applications where complete control of the end-effector is essential. Trajectory planning produces accurate movements of a PR by avoiding Type II singularities. Generally, singularity avoidance is achieved by optimising a geometrical path with a velocity profile considering singular configurations as obstacles. This research presents an algorithm that avoids Type II singularities by modifying the trajectory of a subset of the actuators. The subset of actuators represents the limbs responsible for a Type II singularity, and they are identified by the angle between two Output Twist Screws. The proposed avoidance algorithm does not require optimisation procedures, which reduces the computational cost for offline trajectory planning and makes it suitable for online trajectory planning. The avoidance algorithm is implemented in offline trajectory planning for a pick and place planar PR and a spatial knee rehabilitation PR.

## 1. Introduction

Parallel robots (PRs) are mechanisms where the end-effector (mobile platform) is linked to the base (fixed platform) by at least two open kinematics chains [1, 2]. This closed-chain architecture provides PR with high rigidity and the ability to handle large loads with high accuracy [3]. The main drawback is the existence of configurations inside of workspace where the number of degrees of freedom (DOFs) changes instantaneously, named Type II singularities by Gosselin and Angeles [4]. A Type II singularity is dangerous due to the fact that the end-effector could move despite all actuators being locked. In robotic rehabilitation [5] and collaborative operations [6], the lack of complete control of the PR movements must be solved to keep the user safe. The optimisation of the workspace [7–10], adding actuators [11–14], and trajectory planning [15–18] are some methods to deal with Type II singularities.

∗Corresponding author
*Email addresses:* jopulza@doctor.upv.es (José L. Pulloquinga), raessan2@doctor.upv.es (Rafael J. Escarabajal), giuprog@isa.upv.es (Ángel Valera), mvalles@isa.upv.es (Marina Vallés), vmata@mcm.upv.es (Vicente Mata)

Trajectory planning computes PR motions satisfying robot dynamics and actuator constraints, minimising energy consumption or execution time while avoiding collisions with obstacles such as Type II singularities [16]. The trajectory planning is performed offline, online, or a combination of both. Offline trajectory planning connects two desired PR locations based on prior information about the PR workspace and the regions with Type II singularities [17]. In contrast, online trajectory planning replaces the preceding information about the workspace and singular regions with a partial view of the workspace [16] or sensor feedback [19].

Sen et al. [20] proposed a path planning algorithm for PRs that constrains the potential energy of the PR to avoid singular configurations. In [21], Khoukhi et al. presented a multi-objective dynamic trajectory planning for PRs where a Type II singularity is avoided by maximising the manipulability of the PR. A path planning method by generating singularity-free C-spaces defined in the vicinity of the two configurations to connect is proposed in [22]. Bourbonnais et al. [23] proposed a stochastic cubic spline optimisation that avoids Type II singular regions while generating smooth trajectories for a five bars PR. In [24] Li et al. generate feasible trajectories for a 4-DOF PR using a quintic B-spline considering singular configurations as optimisation constraints. The methods reported in [20–24] are mathematically complex and require considerable effort to select the parameters of the optimisation problem. Thus, Type II singularities avoidance is challenging to implement in real-time applications. Agarwal et al. [25] proposed a task-priority controller where a potential function allows online singularity avoidance that requires reducing one DOF of the PR. Hill et al. [26] applied virtual constraints in a controller to online trajectory generation during a Type II singularity crossing. These methods require setting the control law to track location and avoid singularities simultaneously which increases the complexity of implementation. In [14, 27] singularity avoidance methods based on redundant PRs are analysed.

The closeness to a Type II singularity could be detected by the determinant of the Forward Jacobian matrix [28, 29], motion/force transmission indices [30] or stiffness indices [31]. However, the kinematic chains involved in the singularity are not identified. Based on Screw Theory, Pulloquinga et al. [32] proposed the angle between two instantaneous screw axes from the Output Twist Screws ($\Omega_{i,j}$) to measure the closeness to a Type II singularity. This research verified the capability of the index $\Omega_{i,j}$ to measure the proximity to a singular configuration by an experimental benchmark in a 4-DOF PR prototype. Next, [33] verifies that the minimum $\Omega_{i,j}$ identifies the pair of kinematic chains responsible for the singular configuration in a 4-DOF PR. Then, the minimum angle $\Omega_{i,j}$ was applied to Type II singularity avoidance during offline trajectory planning for a 4-DOF PR [34]. However, Type II singularity avoidance based on the minimum $\Omega_{i,j}$ is limited to a spatial PR, i.e., a general singularity avoidance algorithm is required to include planar PRs. Moreover, online Type II singularity avoidance based on Output Twist Screws (OTSs) has not been analysed.

This paper proposes an algorithm that avoids a Type II singularity by modifying the trajectory of the actuators involved in the singular configuration for planar and spatial PR. The singularity avoidance algorithm works with one trajectory sample at a time. In the vicinity of a Type II singularity, the avoidance algorithm calculates the deviation required in the subset of actuators responsible for the singularity. The actuators responsible for the singularity are identified by the minimum angle between the linear components from two Output Twist Screws ($\Theta_{i,j}$) in the planar PR and by the minimum angle $\Omega_{i,j}$ in the spatial case. Only the trajectory of the actuators identified by the minimum $\Theta_{i,j}$ or $\Omega_{i,j}$ are modified to avoid a Type II singularity, i.e., the reference trajectory requires a minimum modification. For this reason, the proposed algorithm is suitable for applications where the significant changes on the reference trajectory are risky, such as human rehabilitation. The modification in the trajectory of the actuators is calculated using a straight line equation, reducing the diffculty of setting the algorithm proposed. The straight line is defined by the sample time of the trajectory and the velocity of the PR. For an arbitrary sample of time, the proposed algorithm only needs the reference and the current location of the PR to avoid a singularity. In offline trajectory planning, the current location of the PR represents the previous reference trajectory sample. In online trajectory planning, the current location of the PR is measured by a 3D tracking sensor or by solving the Forward kinematics based on encoders in the actuators. Thus, the contribution of this paper is a general Type II singularity avoidance algorithm with low complex calculations and minimal requirements for implementation, suitable for offline and online applications. It is important to emphasise that it is the first time the minimum angle $\Theta_{i,j}$ is applied for singularity avoidance in planar PRs. Therefore, this paper complements the research developed with spatial PRs in [32, 34].

Section 2 presents the mathematical foundations concerned with Type II singularities and their detection by the angles $\Theta_{i,j}$ or $\Omega_{i,j}$. In Section 3, a detailed description of the developed algorithm for Type II singularity avoidance is exposed. Section 4 applies the avoidance algorithm in offline trajectory planning for a five bars PR and discusses the simulation results. Section 5 applies the avoidance algorithm in offline and online trajectory planning for 4-DOF PR for knee rehabilitation. The offline singularity avoidance is verified by measuring the location reached by the actual PR using a 3D tracking system. Then the offline and online trajectory planning results are discussed. Finally, the main conclusions are presented in Section 6.

## 2. Mathematical foundation

### 2.1. Singularities in parallel robots

In a close kinematics mechanism [4], the relationship between the input and output coordinates is defined by a set of constraint equations $\phi$ as follows:

$$\phi(x, q_{ind}) = 0 \qquad (1)$$

where the output $x$ stands for the location and orientation of the end-effector or mobile platform, and the input $q_{ind}$ represents the subset of actuated joints.

Taking time derivatives of (1), the relationship between the outputs and inputs velocities is:

$$J_I \dot{q}_{ind} + J_D \dot{x} = 0 \qquad (2)$$

with $J_I$, $J_D$ as the Inverse and Forward Jacobian matrix, respectively. Both matrices are square matrices ($F \cross F$) for non-redundant Parallel Robots (PRs), where $F$ is the number of DOFs of the mobile platform.

Based on the input-output kinematic relationship, Gosselin and Angeles [4] define three types of singularities:

I The mobile platform of the PR loses mobility in at least one direction where the $J_I$ matrix becomes rank deficient, $\|J_I\| = 0$. Figure 1a shows an example using a five bars PR (5R).

II The mobile platform of the PR gains at least one uncontrollable motion despite all actuators being locked (see Figure 1b), where the $\|J_D\| = 0$.

III Both $J_I$ and $J_D$ become rank deficient. This configuration is only possible for specific values of geometric parameters.

Type II singularities require special attention because they appear within the workspace. Losing control over the mobile platform movements ($\dot{x} \neq \mathbf{0}$) despite the actuators being fixed ($\dot{q}_{ind} = \mathbf{0}$) represents a potential danger for the user or the PR itself. Another drawback of Type II singularities is the increase in the efforts of the actuators because the calculus involves the $J_D$. For PR with $F < 6$, there are constrain singularities quite analogous to Type II singularities [2]. This research is focused on Type II singularities.

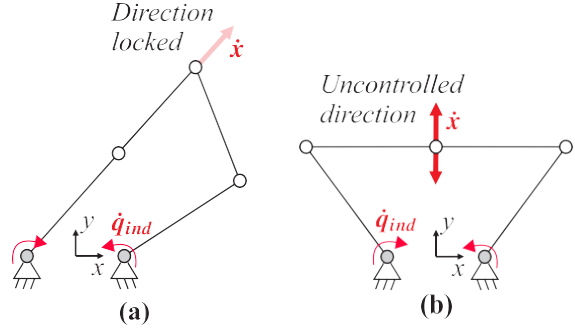

Figure 1: 5R parallel mechanism in singular configuration (a) Type I (b) Type II.

The calculation of $\|J_D\|$ presents a simple numerical method to detect a Type II singularity. However, combining rotational and translational joints produces a $J_D$ matrix with not homogeneous units that make it difficult to measure the closeness to a singularity. Another disadvantage of $\|J_D\|$ is the inability to identify the limbs involved in the Type II singularity.

*2.2. Singularities detection based on Output Twist Screws*

In a non-redundant PR, the motion of the mobile platform is produced by the combined action of $F$ actuators, i.e., the contribution of each actuator is challenging to identify. According to Takeda and Funabashi [35], if all actuators are locked except one, the instantaneous unit motion of the mobile platform is represented by an Output Twist Screw (OTS). Then, the linear and angular velocity of an arbitrary point of the mobile platform $\$_V$ is a linear combination of $F$ unit OTSs $\hat{\$}_O$:

$$\$_V = k_1 \hat{\$}_{O_1} + k_2 \hat{\$}_{O_2} + \ldots + k_F \hat{\$}_{O_F} \qquad (3)$$

where $k_1, k_2 \ldots k_i$ are the amplitude of each OTS.

For an $i = 1 \ldots F$ limb, $\hat{\$}_{O_i}$ is produced by the unit wrench transmitted by the correspondent actuator $\hat{\$}_{T_i}$ while the other actuators apply no work to the mobile platform (see Figure 2). Thus, each $\hat{\$}_{O_i}$ is defined by solving:

$$\hat{\$}_{O_i} \circ \hat{\$}_{T_j} = 0 \; (i, j = 1, 2, \ldots, F, i \neq j) \qquad (4)$$

with

$$\hat{\$}_{O_i} = (\mu_{w_{O_i}}; \mu^*_{v_{O_i}}) \qquad (5)$$

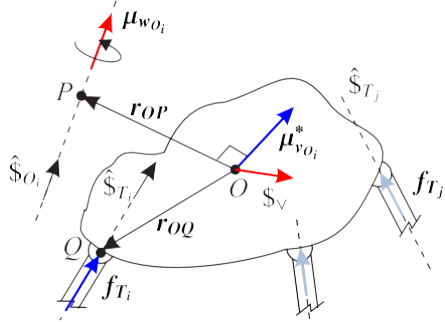

Figure 2: Motion decomposition for a PR.

where ∘ stand for the reciprocal product, $\mu_{w_{O_i}}$ is the instantaneous screw axis, and $\mu^*_{v_{O_i}}$ is the linear component of the $\hat{\$}_{O_i}$.

In [8], Wang et al. proved that for a singular configuration, at least two OTSs are linearly dependent. Then, in a Type II singularity, at least two limbs are contributing to the mobile platform motion in the same direction, i.e.:

$$\hat{\$}_{O_i} = \hat{\$}_{O_j} \; (i \neq j) \quad (6)$$

where $i, j$ identify the limbs under analysis.

From a geometrical point of view, in a singular configuration the linear and angular components of the $\hat{\$}_{O_i}$ and $\hat{\$}_{O_j}$ are parallel. Thus, the parallelism in the linear components $\mu^*_{v_O}$ is measured by the angle $\Theta_{i,j}$, and by the angle $\Omega_{i,j}$ for the angular components $\mu_{w_O}$, as follows:

$$\Theta_{i,j} = \arccos(\mu^*_{v_{O_i}} \cdot \mu^*_{v_{O_j}}) \; (i,j = 1,2,\ldots,F, i \neq j) \quad (7)$$

$$\Omega_{i,j} = \arccos(\mu_{w_{O_i}} \cdot \mu_{w_{O_j}}) \; (i,j = 1,2,\ldots,F, i \neq j) \quad (8)$$

In a Type II singularity $\Theta_{i,j} = \Omega_{i,j} = 0$, where the limbs $i, j$ are responsible for the configuration because they contribute to the mobile platform motion in the same direction. The $\Theta_{i,j}$ and $\Omega_{i,j}$ have the advantage of having physical meaning because they are angular magnitudes as opposed to the $\|J_D\|$. Examples of calculating the angles $\Theta_{i,j}$ and $\Omega_{i,j}$ are presented at the beginning of Section 4 and Section 5, respectively.

For a planar PR ($F = 2$), the closeness to a Type II singularity is measured by decreasing $\Theta_{i,j}$. It is because the $\mu_{w_{O_i}}$ and $\mu_{w_{O_j}}$ are perpendicular to the plane maintaining $\Omega_{i,j} = 0$.

For a spatial PR ($F > 2$), the combination of pairs of limbs in expressions (7) and (8) generates several possible angles $\Theta_{i,j}$ and $\Omega_{i,j}$, respectively. However, the parallelism between two $\hat{\$}_O$ is the minimum condition to reach a singular configuration. Then, the proximity to a Type II singularity is defined by the minimum angles $\Theta_{i,j}$ and $\Omega_{i,j}$. However, in (5), the $\mu^*_{v_{O_i}}$ can be rewritten with respect to the $\mu_{w_{O_i}}$ as follows:

$$\mu^*_{v_{O_i}} = h\mu_{w_{O_i}} + r_{OP} \times \mu_{w_{O_i}} \quad (9)$$

where $h$ is the screw's pitch and $r_{OP}$ is the minimal distance between the selected point of the mobile platform $O$ and $\mu_{w_{O_i}}$, see Figure 2.

Rewriting the cross product with matrix multiplication (9) becomes:

$$\mu^*_{vo_i} = (H + \tilde{R})\mu_{wo_i} \tag{10}$$

with the matrix $H$ and the skew-symmetric matrix $\tilde{R}$ defined as:

$$H = \begin{bmatrix} h & 0 & 0 \\ 0 & h & 0 \\ 0 & 0 & h \end{bmatrix} \quad \tilde{R} = \begin{bmatrix} 0 & -r_{OP_z} & r_{OP_y} \\ r_{OP_z} & 0 & -r_{OP_x} \\ -r_{OP_y} & r_{OP_x} & 0 \end{bmatrix} \tag{11}$$

According to (10), if $H$ and $\tilde{R}$ are null matrices, $\mu^*_{vo_i} = 0$ even though $\mu_{wo_i} \neq 0$, i.e., the limb $i$ contributes with pure angular motion. Then, there are arbitrary non-singular configurations with $\Theta_{i,j} = 0$ and $\Omega_{i,j} \neq 0$ because the $\mu^*_{vo_i}$ and $\mu^*_{vo_j}$ could disappear or become parallel. In contrast, the angle $\Omega_{i,j} = 0$ if and only if the $\mu_{wo_i}$ and $\mu_{wo_j}$ become parallel. Therefore, in a spatial PR, the closeness to a Type II singularity depends mainly on decreasing the minimum angle $\Omega_{i,j}$. The effectiveness of the minimum angle $\Omega_{i,j}$ as a Type II singularity proximity detector was proved by an experimental benchmark in [32]. In that work, the minimum angle $\Theta_{i,j}$ was used to verify that the PR reaches a singular configuration. An example of the Type II singularity proximity detection using the minimum angle $\Omega_{i,j}$ is presented in Section 5.

The minimum angle $\Omega_{i,j}$ has been used to avoid singularities during an offline trajectory planning in a 4-DOF PR [34]. However, the index $\Theta_{i,j}$ has not been used for Type II singularities avoidance in planar PRs. This paper presents a novel Type II singularity avoidance algorithm for general purposes based on the minimum angles $\Omega_{i,j}$ and $\Theta_{i,j}$. The avoidance algorithm uses $\Theta_{i,j}$ to detect the proximity to a singularity in the planar PR and $\Omega_{i,j}$ for the spatial case. Moreover, the execution time of the avoidance algorithm is measured to verify the low computational cost of calculating the minimum angles $\Omega_{i,j}$ and $\Theta_{i,j}$.

## 3. Type II Singularity Avoidance Algorithm

For a discretised instant time, the reference pose of the mobile platform $x_r$ and the actual pose reached by the PR $x_m$ are the inputs required by the avoidance algorithm. Each discretised instant time, the inputs $x_r$ and $x_m$ are taken simultaneously. The Inverse Kinematic model calculates the correspondent reference location in joint space $q_{ind_r}$. Figure 3 shows an overview of the algorithm proposed.

The avoidance algorithm first takes $x_r$ to calculate all $\hat{\$}_{O_i}$ ($i = 1\ldots F$), and calculates the arrays $v\Theta_r$ and $v\Omega_r$ that store all possible angles $\Theta_{i,j}$ and $\Omega_{i,j}$ ($i,j = 1\ldots F, i \neq j$), respectively. Next, $a_r$ calculates the minimum element in $v\Theta_r$ for planar PRs or in $v\Omega_r$ for a spatial case. Analogously, the $v\Theta_m$, $v\Omega_m$, $a_m$ are calculated for the measured $x_m$, and $i_{ch}$ stores the two limbs identified by $a_m$.

The proposed algorithm calculates a non-singular desired location $q_{ind_d}$ by using:

$$q_{ind_d} = q_{ind_r} + v_d t_s \Delta\iota \tag{12}$$

$v_d$ and $t_s$ are the constant velocity for singularity avoidance and the sample time for the trajectory discretisation, respectively. $\Delta\iota$ is an integer accumulator ($F\times 1$) for the deviation required in the actuators. The non-singular desired reference in configuration space $x_d$ could be calculated by solving the Forward Kinematics problem for $q_{ind_d}$.

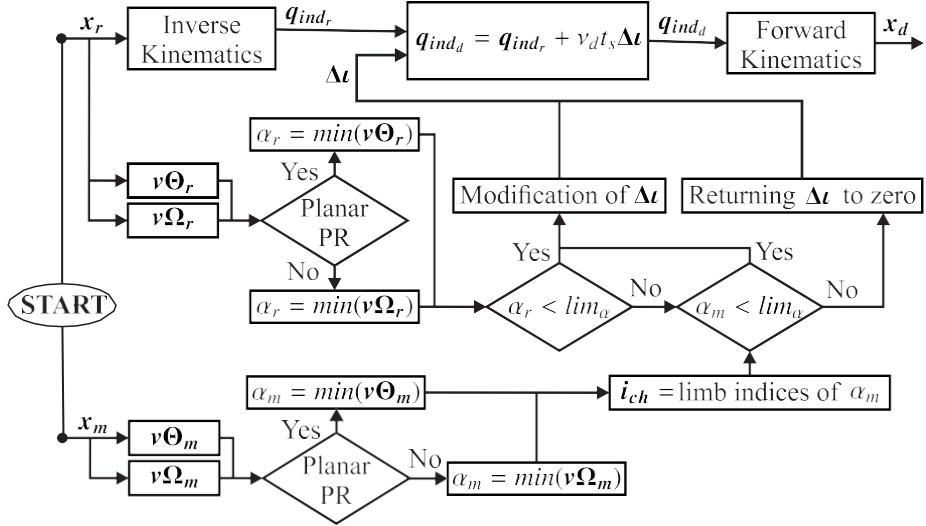

Figure 3: Block diagram for Type II singularity avoidance algorithm.

The proposed algorithm starts with $q_{ind_d} = q_{ind_r}$, i.e., $\Delta\iota = 0$. If $\alpha_r$ is below a predefined threshold ($lim_\alpha$), the $\Delta\iota$ is modified because the reference $x_r$ is proximal to a Type II singularity. Two rows of $\Delta\iota$ are modified by one at each discretised instant time until the $q_{ind_d}$ drives the PR to a non-singular configuration $x_m$, i.e., $\alpha_m > lim_\alpha$. With $\alpha_m > lim_\alpha$, if the reference $x_r$ becomes non-singular ($\alpha_r > lim_\alpha$), $\Delta\iota$ must return progressively to zero because a Type II singularity has been avoided. Figure 4a shows a detailed process of modifying $\Delta\iota$ when the PR is proximal to a Type II singularity. In contrast, Figure 4b explains how $\Delta\iota$ returns to zero after Type II singularity avoidance.

In Figure 4a, the $\Delta\iota$ is modified by one only in the two rows corresponding to the elements in $i_{ch}$. Each row, identified by $i_{ch}$, could hold, increase or decrease, generating eight possible modifications of $\Delta\iota$. Considering that an actuator could stop (0), go forwards (1) or go backwards (-1), the eight possible modifications are grouped as columns of the matrix $M_{av}$, see (13). For each $k$ possible modification of $\Delta\iota$, the $q_{ind_d}$ and $\alpha_d$ are calculated. The index $\alpha_d$ is the element in $v\Theta_d$ or in $v\Omega_d$ for the limbs $i_{ch}$. The $v\Theta_d$ and $v\Omega_d$ allocate all possible angles $\Theta_{i,j}$ and $\Omega_{i,j}$ for the location $x_d$. The best modification of $\Delta\iota$ is selected to generate a feasible $q_{ind_d}$ that steps the PR away from the singularity, i.e., $\alpha_d > \alpha_m$. The deviation on the reference trajectory continues in successive iterations until $x_m$ becomes non-singular, $\alpha_m > lim_\alpha$.

$$M_{av} = \begin{bmatrix} 1 & -1 & 1 & -1 & 1 & -1 & 0 & 0 \\ 1 & -1 & -1 & 1 & 0 & 0 & 1 & -1 \end{bmatrix} \tag{13}$$

In the case of Figure 4b, the two rows of $\Delta\iota$ with the maximum value ($max\Delta$) are identified and saved in $i_{re}$. The $\Delta\iota$ is modified in the rows $i_{re}$ according to the columns of $M_{av}$, generating eight new possible $\Delta\iota$. The proper $\Delta\iota$ is selected to ensure a feasible $q_{ind_d}$ and to decrease the absolute value of $\Delta\iota$. The returning procedure continues until $\Delta\iota = 0$.

Figure 3 shows that the proposed algorithm does not require complex optimisation problems, making it suitable for offline trajectory planning and combining with controllers for online Type

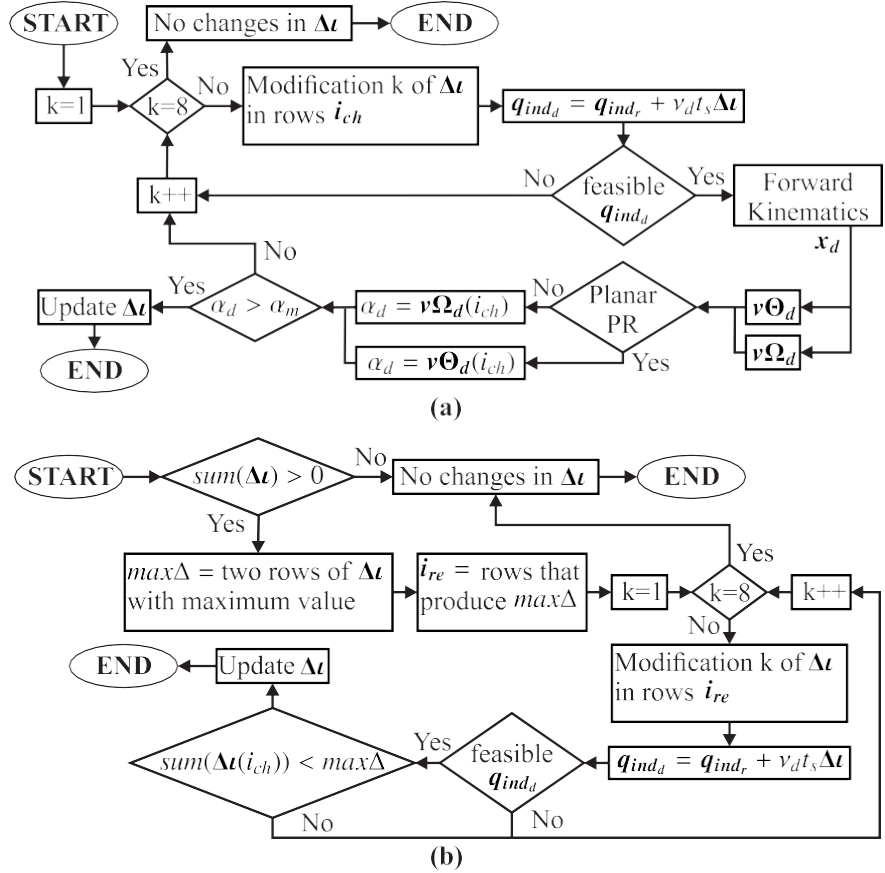

Figure 4: Operation on $\Delta\iota$ :(a) modification (b) return to zero.

II singularity avoidance. For online applications, the algorithm is implemented by setting $t_s$ as the controller sample time and $v_d$ as the average feasible velocity for the PR under analysis. $x_m$ could be measured by a 3D tracking system or by estimation based on embedded encoders. In offline trajectory planning, the avoidance algorithm is applied at each sample of a discrete reference trajectory of $N_{ptos}$ samples ($N_{ptos} > 1$). The reference trajectory could have different shapes, including a linear interpolation between two different desired points. The avoidance parameter $t_s$ is set to the sample time of the reference trajectory while $v_d$ is the average of the velocity reference trajectory. For offline trajectory planning, the measured signal $x_m$ is initialised with $x_r$, and for the consequent iterations, $x_d$ is used as a feedback signal, i.e., $x_m = x_d$.

The value of $lim_\alpha$ is calculated previously to implement the Type II singularity avoidance algorithm. First, the fundamental movements of the PR are combined to generate a set of trajectories that approach a Type II singularity from different non-singular configurations. The application under study defines the fundamental movements of PR. Next, 70 % of the trajectories are executed until the PR under study loses control of the mobile platform. During the execution of trajectories, the pose reached by the PR must be measured. A set of minimum $\Theta_{i,j}$ and minimum $\Omega_{i,j}$ is calculated based on the pose of the PR measured at each configuration of the trajectories executed. Subsequently, the $lim_\alpha$ is defined as the average of the set of minimum $\Theta_{i,j}$ for planar PR or the set of minimum $\Omega_{i,j}$ for the spatial case. Next, the remaining 30 % of trajectories are executed until the PR reaches the $lim_\alpha$. If the PR holds control over the mobile platform with the actuators locked, the setting of $lim_\alpha$ is finished. A detailed explanation of the experimental procedure to set the $lim_\alpha$ is presented in [32].

The Type II singularity avoidance algorithm presents the following advantages:

- At each discretised instant time, only two actuators require modifications for singularity avoidance.
- The avoidance parameters $v_d$ and $t_s$ are set directly by the PR application.
- Low computational cost due to the absence of optimisation functions. The execution time of the avoidance algorithm is compared with existing algorithms in Sections 4 and 5.

The proposed avoidance algorithm is designed to modify the trajectory of two actuators because, in the closeness of a Type II singularity, at least two OTSs are parallel. The case of three OTSs aligned appears when the PR already reaches the singular configuration [32]. However, if three OTSs become parallel in closeness to a singular configuration, the algorithm modifies the trajectory of the actuators by pairs. For example, consider a spatial PR at an arbitrary instant time. If the $\Omega_{1,2}$ and $\Omega_{1,3}$ are the minimum elements in $\alpha_m$ the avoidance algorithm modifies the trajectory for actuators 1 and 2. Assuming that the $\Omega_{1,2}$ has increased for the next discretised instant time, the proposed algorithm modifies the trajectory on actuators 1 and 3.

It is important to emphasise that the proposed avoidance algorithm is limited to Type II singularities. It is because the minimum angles $\Omega_{i,j}$ and $\Theta_{i,j}$ measure when two actuators transmit motion to the end-effector in the same direction, i.e., they detect the loss of controlled DOFs. The proposed avoidance algorithm is unsuitable for Type I singularities because, in this configuration, the PR retains control over all DOFs.

## 4. Case of study: 5R parallel mechanism

The 5R parallel mechanism is a planar PR with 2-DOF used for positioning a point P on a defined plane (see Figure 5). Point P is connected to the base by two limbs each of which consists of two links. The mechanism is named 5R because the links are connected by revolute joints where the two joints connected to the base are actuated.

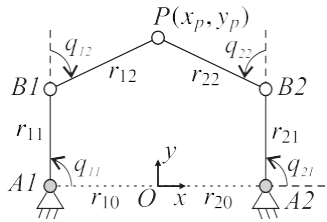

Figure 5: Simplified representation of 5R parallel mechanism.

The kinematic model of the 5R mechanism is shown in Figure 5 where $r_{11}$, $r_{12}$, $r_{21}$ and $r_{22}$ are the length of the links. $r_{10}$ and $r_{20}$ are the horizontal distance to the connecting points

A1 and A2 measured from O in the fixed frame {$O\text{-}xyz$} respectively. The active joints are defined as $q_{11}$, $q_{21}$, while the passive joints are $q_{12}$, $q_{22}$. In this research, the 5R mechanism has a symmetrical architecture where the inverse kinematic model considers the working mode $+$ [36]. The geometrical parameters of the 5R mechanism are shown in Table 1.

Table 1: Geometrical parameters of the 5R mechanism in meters

| $r_{10}, r_{20}$ (m) | $r_{11}, r_{21}$ (m) | $r_{12}, r_{22}$ (m) |
|---|---|---|
| 0.04 | 0.06 | 0.05 |

The movement of the 5R mechanism is divided into $\hat{\$}_{O_1}$ and $\hat{\$}_{O_2}$, which by using (7) and (8) define the indices $\Theta_{1,2}$ and $\Omega_{1,2}$, respectively. In the 5R mechanism, $\Omega_{1,2} = 0$ because the movement takes place in the plane $xy$ remaining the instantaneous screw axis in the $z$ axis.

At point P, the $\hat{\$}_{O_1}$ represents the contribution of limb 1 to the motion of the point P, i.e.:

$$\hat{\$}_{O_1} = (\begin{matrix} 0 & 0 & 1 \end{matrix}\ ;\ \begin{matrix} v_x & v_y & 0 \end{matrix}) \tag{14}$$

Considering the reciprocal product property and the unitary norm of the linear motion, the components $v_x$, $v_y$ are calculated by solving the non-linear system:

$$\begin{aligned} \hat{\$}_{O_1} \circ \hat{\$}_{T_2} &= 0 \\ v_x^2 + v_y^2 &= 1 \end{aligned} \tag{15}$$

Due to the rotational actuator on limb 2 ($q_{21}$) transmitting pure force to point P, the $\hat{\$}_{T_2}$ is:

$$\hat{\$}_{T_2} = (\begin{matrix} f_{T_2}^T \end{matrix}\ ;\ \begin{matrix} 0 & 0 & 0 \end{matrix}) \tag{16}$$

$f_{T_2}$ stands for the unitary force vector in the direction of the link B2P represented on the fixed frame {$O\text{-}xyz$}. Considering two moving frames attached to the links A2B2 and B2P, the $f_{T_2}$ is calculated as follow:

$$f_{T_2} = R_1\ {}^1R_2\ [\begin{matrix} 1 & 0 & 0 \end{matrix}]^T \tag{17}$$

where the rotation matrix $R_1$ and ${}^1R_2$ are defined as:

$$R_1 = \begin{bmatrix} \cos q_{21} & -\sin q_{21} & 0 \\ \sin q_{21} & \cos q_{21} & 0 \\ 0 & 0 & 1 \end{bmatrix} \quad {}^1R_2 = \begin{bmatrix} \cos q_{22} & -\sin q_{22} & 0 \\ \sin q_{22} & \cos q_{22} & 0 \\ 0 & 0 & 1 \end{bmatrix} \tag{18}$$

Developing (17) with (18), $f_{T_2}$ is:

$$f_{T_2} = \begin{bmatrix} \cos q_{21} \cos q_{22} - \sin q_{21} \sin q_{22} & \sin q_{21} \cos q_{22} + \cos q_{21} \sin q_{22} & 0 \end{bmatrix}^T \tag{19}$$

The $\hat{\$}_{O_2}$ is calculated solving (15) replacing $\hat{\$}_{T_2}$ with the pure force screw $\hat{\$}_{T_1}$ as:

$$\hat{\$}_{T_1} = (\begin{matrix} f_{T_1}^T \end{matrix}\ ;\ \begin{matrix} 0 & 0 & 0 \end{matrix}) \tag{20}$$

with the force $f_{T_1}$ in the direction of the link B1P defined as follow:

$$f_{T_1} = \begin{bmatrix} \cos q_{11} \cos q_{12} - \sin q_{11} \sin q_{12} & \sin q_{11} \cos q_{12} + \cos q_{11} \sin q_{12} & 0 \end{bmatrix}^T \tag{21}$$

*4.1. Offline planning results*

The offline trajectory planning starts with generating a constant velocity trajectory on the plane $xy$ with a Type II singularity in the middle. The description of the trajectory in configuration space is shown in Table 2. Next, the singularity avoidance algorithm is applied to the original trajectory to obtain the non-singular version of the trajectory. The 5R mechanism is driven by an Arduino Uno board at a rate of 200 Hz with a maximum working velocity of 29 °/s. Therefore, the singularity avoidance algorithm is set to $t_s = 20\ ms$ and $v_d = 0.5\ rad/s$.

Table 2: Description of the trajectory for offline planning in the 5R mechanism.

| Location | $x_p$ (m) | $y_p$ (m) | time (s) |
|---|---|---|---|
| Start | 0 | 0.09 | 0 |
| Singularity | -0.03 | 0.05 | 2 |
| End | 0 | 0.09 | 4 |

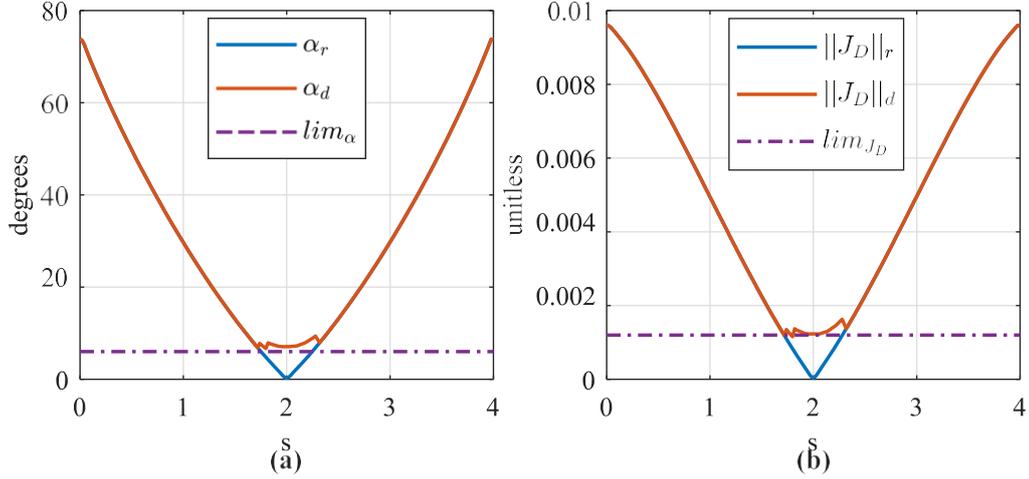

Figure 6: Type II singularity proximity in 5R mechanism (a) $\alpha$ (b) $\|J_D\|$.

For 5R mechanism $\alpha = \Theta_{1,2}$, Figure 6a shows that the non-singular trajectory holds $\alpha_d$ greater than $lim_\alpha$ while the original trajectory reaches a singularity at 2 s ($\alpha_r = 0$). The threshold for avoidance is set to 6 ° ($lim_\alpha = 0.1047\ rad$) based on several simulation results. The Type II singularity avoidance is verified by analysing the $\|J_D\|$ of the original ($\|J_D\|_r$) and the non-singular trajectory ($\|J_D\|_d$). Figure 6b shows that $\|J_D\|_d$ always differs from zero while $\|J_D\|_r$ reaches a Type II singularity at 2 s. In this research, the $\|J_D\|$ is not deeply analysed. However, [37] shows the expression of $\|J_D\|$ considering $x = [x_p\ y_p]^T$.

Figure 7 shows the original ($*_r$) and the non-singular trajectory ($*_d$) generated for the actuator on limbs 1 ($* = q_{11}$) and 2 ($* = q_{21}$).

Figure 7 shows that the proposed algorithm requires a maximum deviation of 1.2 ° in the joint trajectories to avoid a Type II singularity. The offline trajectory planning was performed in MATLAB on a desktop PC with 32 GB of RAM. Table 3 compares the proposed algorithm with a constrained multi-objective (CMO) algorithm and roadmap (C-space) algorithm. In Table 3, $t_I$ represents the average elapsed time in each iteration, and $t_T$ is the total trajectory time. The results for the CMO algorithm were taken from [21] for offline trajectory planning in a planar PR. The results for the C-space algorithm were calculated based on the information reported in [22] for offline path planning in 3-RRR mechanism. According to Table 3, the proposed avoidance algorithm is the fastest during offline trajectory planning, with an average elapsed time of 0.5 ms. Thus, the Type II singularity avoidance algorithm is suitable for offline trajectory planning with a minimum modification of the original trajectory and low computational cost.

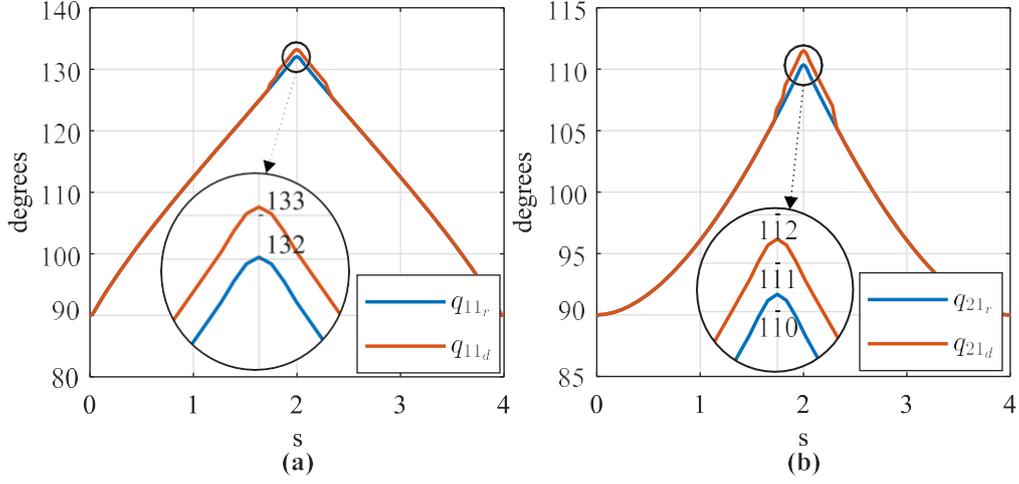

Figure 7: Trajectory generated for (a) $q_{11}$ (b) $q_{21}$ in the 5R mechanism.

Table 3: Comparative results of avoidance algorithm for offline trajectory planning in planar PRs.

| Algorithm | $t_I$ (s) | $t_T$ (s) | Processor |
|---|---|---|---|
| Proposed | 0.0005 | 4 | Intel Core i7 3.7 GHz |
| CMO [21] | 1.2620 | 9.5 | unspecified |
| C-space [22] | 0.0756 | 1 | Intel Core i7 2.66 GHz |

## 5. Case of study: 3UPS+RPU PR

The 3UPS+RPU PR is a 4-DOF mechanism for knee rehabilitation and diagnosis purposes. Figure 8 shows the actual prototype and its kinematic model. The 4-DOF of the PR are two translational movements ($x_m$, $z_m$) in the tibiofemoral plane, one rotation ($\psi$) around the coronal plane and one rotation ($\theta$) around the tibiofemoral plane [38]. These four DOFs are controlled by four linear actuators represented by $q_{13}$, $q_{23}$, $q_{33}$ and $q_{42}$, see Figure 8b. The designation 3UPS+RPU refers to the three external limbs with UPS configuration and the central one with RPU configuration (Figure 8b). The letters R, U, S and P represent revolute, universal, spherical and prismatic joints, respectively, and "_" identifies the actuated joint.

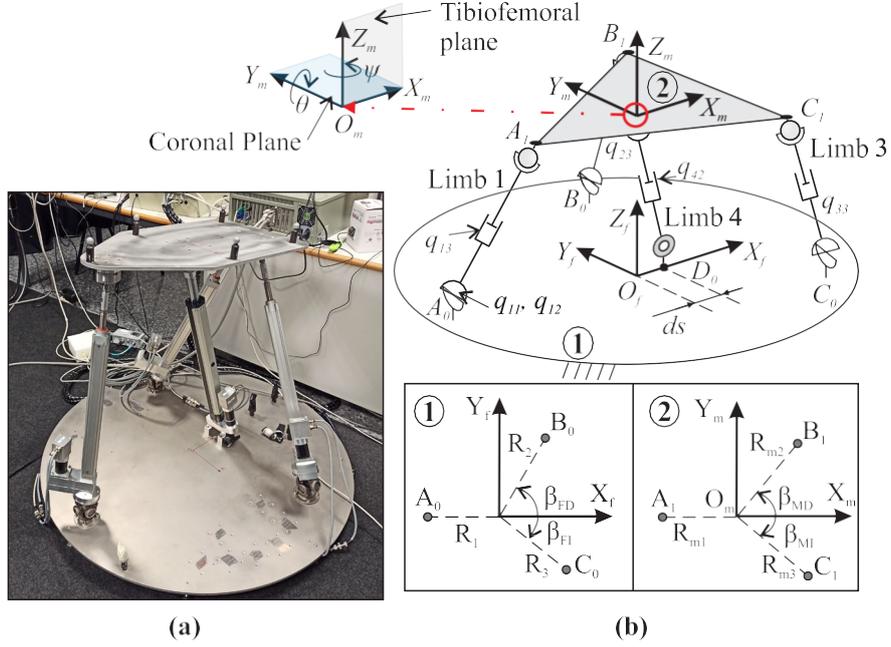

Figure 8: 3UPS+RPU PR: (a) actual prototype (b) simplified representation.

The four points that connect the limbs with the fixed platform ($A_0, \ldots, D_0$) are located by the geometric variables $R_1$, $R_2$, $R_3$, $\beta_{FD}$, $\beta_{FI}$ and $ds$. The four points that connect the limbs with the mobile platform ($A_1$, $B_1$, $C_1$, $O_m$) are located by the geometric variables $R_{m1}$, $R_{m2}$, $R_{m3}$, $\beta_{MD}$ and $\beta_{MI}$. The geometric parameters on the fixed platform are measured with respect to the fixed frame $\{O_f - X_f Y_f Z_f\}$ while the parameters on the mobile platform with respect to the moving frame $\{O_m - X_m Y_m Z_m\}$. The geometrical parameters of the PR under analysis are shown in Table 4.

Table 4: Geometrical parameters of the 3UPS+RPU PR

| $R_1$, $R_2$, $R_3$ (m) | $\beta_{FD}$ (°) | $\beta_{FI}$ (°) | $ds$ (m) | $R_{m1}$, $R_{m2}$, $R_{m3}$ (m) | $\beta_{MD}$ (°) | $\beta_{MI}$ (°) |
|---|---|---|---|---|---|---|
| 0.4 | 90 | 45 | 0.15 | 0.3 | 50 | 90 |

The movement of the 3UPS+RPU PR is divided into $\hat{\$}_{O_1}$, $\hat{\$}_{O_2}$, $\hat{\$}_{O_3}$ and $\hat{\$}_{O_4}$ that using (8) define the six indices $\Omega_{1,2}$ $\Omega_{1,3}$, $\Omega_{1,4}$, $\Omega_{2,3}$, $\Omega_{2,4}$ and $\Omega_{3,4}$. Analogously, (7) generates six indices $\Theta_{1,2}, \ldots, \Theta_{3,4}$. Section 2.2 mentioned that for a spatial PR, the minimum angle $\Theta_{i,j} = 0$ in some non-singular configurations. Figure 9a shows the angles $\Theta_{1,3}$ and $\Theta_{3,4}$ during a trajectory that goes from a non-singular configuration to a Type II singularity in the 4-DOF PR. Figure 9b shows the angles $\Omega_{1,3}$ and $\Omega_{3,4}$ for the same trajectory to a Type II singularity. This figure shows that the minimum angle $\Theta_{i,j}$ ($\Theta_{1,3}$) disappears for non-singular configurations, only the minimum angle $\Omega_{i,j}$ ($\Omega_{3,4}$) has a continuous decrement in the proximity to singularity. Therefore, the closeness to a Type II singularity is measured by the minimum angle $\Omega_{i,j}$. The constraint singularities are not analysed because the 3UPS+RPU PR has mechanical constraints for linear motion on the $Y_f$ axis and the angular motion around the mobile axis $X_m$, i.e., the constraint wrench screws are always linearly independent.

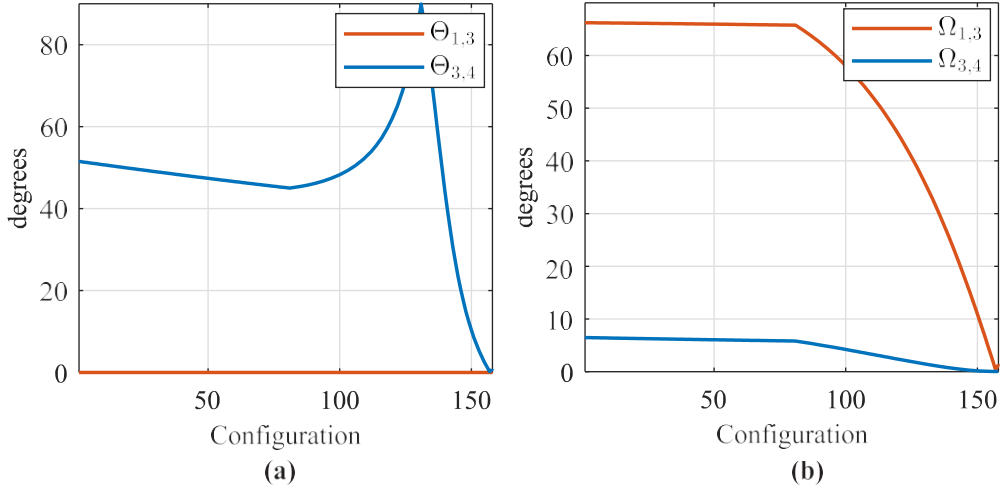

Figure 9: Angles (a) $\Theta_{i,j}$ (b) $\Omega_{i,j}$ during a trajectory approaching a Type II singularity.

The contribution to the motion of the limb 1 $\hat{\$}_{O_1}$ in point $O_m$ with respect to the fixed frame is:

$$\hat{\$}_{O_1} = \begin{pmatrix} w_x & w_y & w_z & ; & v_x & 0 & v_z \end{pmatrix} \quad (22)$$

For an arbitrary point on the mobile platform of the 3U<u>P</u>S+R<u>P</u>U PR, $w_x$ is related to $w_z$ as follows:

$$w_x = w_z \tan\Theta \quad (23)$$

The expression in (23) is determined by taking the time derivatives of the rotation matrix between the moving and fixed frame $^f R_m$. Considering the Euler Y-Z' angle convention $^f R_m$ is:

$$^f R_m = \begin{bmatrix} \cos\theta\cos\psi & -\cos\theta\sin\psi & \sin\theta \\ \sin\psi & \cos\psi & 0 \\ -\sin\theta\cos\psi & \sin\theta\sin\psi & \cos\theta \end{bmatrix} \quad (24)$$

Based on the reciprocal product, the relationship (23) and the unitary norm of the instantaneous screw axis, $w_x$, $w_y$, $w_z$, $v_x$ and $v_z$ are calculated by solving the non-linear system:

$$\begin{aligned} \hat{\$}_{O_1} \circ \hat{\$}_{T_2} &= 0 \\ \hat{\$}_{O_1} \circ \hat{\$}_{T_3} &= 0 \\ \hat{\$}_{O_1} \circ \hat{\$}_{T_4} &= 0 \\ w_x - w_z \tan\Theta &= 0 \\ w_x^2 + w_y^2 + w_z^2 &= 1 \end{aligned} \quad (25)$$

In this case, the actuators transmit force and moment to the mobile platform because the external limbs are not connected directly to the point $O_m$. Thus the transmission wrench $\hat{\$}_{T_1}$, $\hat{\$}_{T_2}$ and $\hat{\$}_{T_3}$ are defined as:

$$\hat{\$}_{T_1} = \begin{pmatrix} f_{T_1}{}^T & ; & (ro_{A_m\,1} \times f_{T_1})^T \end{pmatrix} \tag{26}$$

$$\hat{\$}_{T_2} = \begin{pmatrix} f_{T_2}{}^T & ; & (ro_{B_m\,1} \times f_{T_2})^T \end{pmatrix} \tag{27}$$

$$\hat{\$}_{T_3} = \begin{pmatrix} f_{T_3}{}^T & ; & (ro_{m C_1} \times f_{T_3})^T \end{pmatrix} \tag{28}$$

where $ro_{m A_1}$, $ro_{m B_1}$ and $ro_{m C_1}$ stand for the location vector between the point $O_m$ and the vertices $A_1$, $B_1$ and $C_1$, respectively.

The central limb is connected to the point $O_m$, so the $\hat{\$}_{T_4}$ is:

$$\hat{\$}_{T_4} = \begin{pmatrix} f_{T_4}{}^T & ; & 0 \; 0 \; 0 \end{pmatrix} \tag{29}$$

$f_T$ is the unit vector of the force applied by each actuator with respect to the fixed frame. The direction of the $f_{T_1}$, $f_{T_2}$ and $f_{T_3}$ depends on the universal joint that connects the external limbs with the fixed platform, see Figure 8b. The direction of the $f_{T_4}$ depends on the revolute joint that connects the central limb with the fixed frame.

The universal joint in limb 1 is represented by two orthogonal rotations $q_{11}$ and $q_{12}$ (Figure 8b). Thus, the $f_{T_1}$ is defined as follows:

$$f_{T_1} = R_1\,{}^1R_2 \begin{bmatrix} 0 & 0 & 1 \end{bmatrix}^T \tag{30}$$

with the rotation matrix $R_1$ and ${}^1R_2$ as:

$$R_1 = \begin{bmatrix} \cos q_{11} & 0 & -\sin q_{11} \\ 0 & 1 & 0 \\ \sin q_{11} & 0 & \cos q_{11} \end{bmatrix} \quad {}^1R_2 = \begin{bmatrix} \cos q_{12} & 0 & \sin q_{12} \\ \sin q_{12} & 0 & -\cos q_{12} \\ 0 & 1 & 0 \end{bmatrix} \tag{31}$$

Developing (30) with (31), $f_{T_1}$ is:

$$f_{T_1} = \begin{bmatrix} \cos q_{11} \sin q_{12} & -\cos q_{12} & \sin q_{11} \sin q_{12} \end{bmatrix}^T \tag{32}$$

The $f_{T_2}$ is equal to the (32) with $q_{21}$ and $q_{22}$ instead of $q_{11}$ and $q_{12}$, respectively. The same replacement is performed to $f_{T_3}$, considering $q_{31}$ and $q_{32}$.

In the central limb, the $f_{T_4}$ is defined by the orientation of the revolute joint ($q_{41}$) as follows:

$$f_{T_4} = \begin{bmatrix} -\sin q_{41} & 0 & \cos q_{41} \end{bmatrix}^T \tag{33}$$

The $\hat{\$}_{O_2}$ is calculated based on (25) by modifying the reciprocal product as follows:

$$\begin{aligned} \hat{\$}_{O_2} \circ \hat{\$}_{T_1} &= 0 \\ \hat{\$}_{O_2} \circ \hat{\$}_{T_3} &= 0 \\ \hat{\$}_{O_2} \circ \hat{\$}_{T_4} &= 0 \\ w_x - w_z \tan\Theta &= 0 \\ w_x^2 + w_y^2 + w_z^2 &= 1 \end{aligned} \tag{34}$$

The $\hat{\$}_{O_3}$ and $\hat{\$}_{O_4}$ are calculated with the analogous process of removing the reciprocal product related to $\hat{\$}_{T_3}$ and $\hat{\$}_{T_4}$, respectively.

## 5.1. Offline and online planning results

For offline trajectory planning, the original trajectory represents a knee rehabilitation exercise, specifically a hip flexion, where a singular configuration arises halfway. The description of the trajectory for hip flexion in configuration space is shown in Table 5. Then, the proposed Type II singularity avoidance algorithm is applied to generate a new non-singular trajectory. The actual 3U$\underline{P}$S+R$\underline{P}$U PR is driven by a PID controller in Robot Operating System 2 (ROS2) using the C++ programming language at a rate of 100 Hz. The maximum working velocity is 0.01 $m/s$ because the actual PR works in knee rehabilitation. For these reasons, the singularity avoidance algorithm is adjusted to $t_s = 10\ ms$ and $v_d = 0.01\ m/s$.

Table 5: Description of the trajectory for offline planning in the 4-DOF PR.

| Location | $x_m$ (m) | $z_m$ (m) | $\theta$ (degrees) | $\psi$ (degrees) | time (s) |
|---|---|---|---|---|---|
| Start | 0.038 | 0.640 | 1.14 | 3.64 | 0 |
| Singularity | 0.016 | 0.707 | 8.619 | 18.15 | 12.76 |
| End | 0.038 | 0.640 | 1.14 | 3.64 | 40.53 |

For the 3U$\underline{P}$S+R$\underline{P}$U PR, $\alpha$ is calculated as the minimum element in $v\Omega = [\Omega_{1,2}\ \Omega_{1,3}\ \ldots\ \Omega_{3,4}]$. For the hip flexion trajectory under analysis $\alpha = \Omega_{3,4}$. Figure 10a verifies that the non-singularity trajectory holds $\alpha_d > lim_\alpha$ and the original trajectory decreases $\alpha_r$ under $lim_\alpha$. After several tests in the actual PR under analysis, the threshold for avoidance is set to 2° ($lim_\alpha = 0.0349\ rad$). In addition, Figure 10b shows that $\|J_D\|$ for the non-singular trajectory ($\|J_D\|_d$) is farther from zero than the original trajectory ($\|J_D\|_r$). These results verify the effectiveness of the Type II singularity avoidance algorithm. In this research, the $\|J_D\|$ is only used for verifying the Type II singularity closeness. The reader could review [32] for the detailed expression of $\|J_D\|$ with $x = [x_m\ z_m\ \theta\ \psi]^T$.

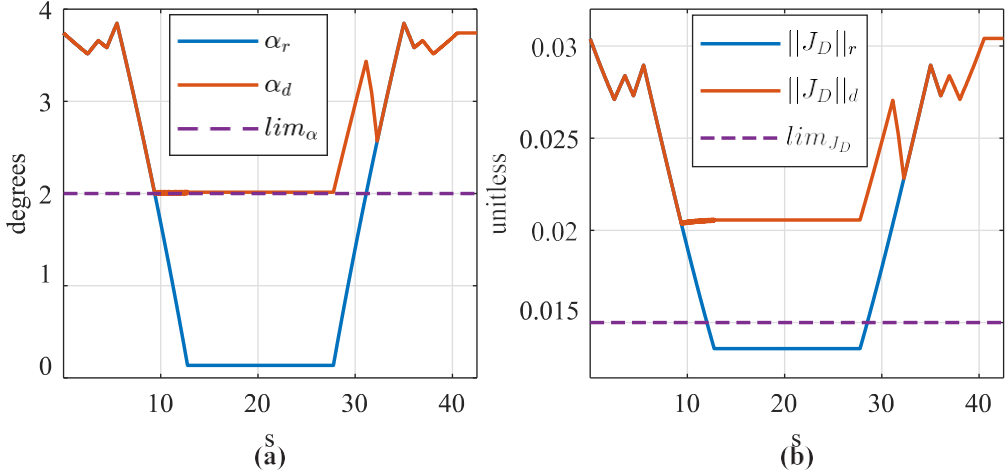

Figure 10: Type II singularity proximity in 3U$\underline{P}$S+R$\underline{P}$U PR (a) $\alpha$ (b) $\|J_D\|$.

The avoidance algorithm deviates the original trajectory of the linear actuators $q_{33}$ and $q_{42}$ to generate a non-singular trajectory because limbs 3 and 4 are involved in the Type II singularity ($\alpha = \Omega_{3,4}$). Figure 11a presents the original trajectory for actuators 3 ($q_{33_r}$) and the non-singular trajectory $q_{33_d}$ generated by the proposed algorithm. Figure 11b presents the original ($q_{42_r}$) and non-singular ($q_{42_d}$) trajectory for actuator 4. In this case, the maximum $\Delta l$ was $[0\ 0\ 56.57]^T$. Figure 11 verifies that the proposed algorithm introduces a smooth deviation in the actuators 3 and 4 to avoid a Type II singularity with a maximum modification of 6 mm.

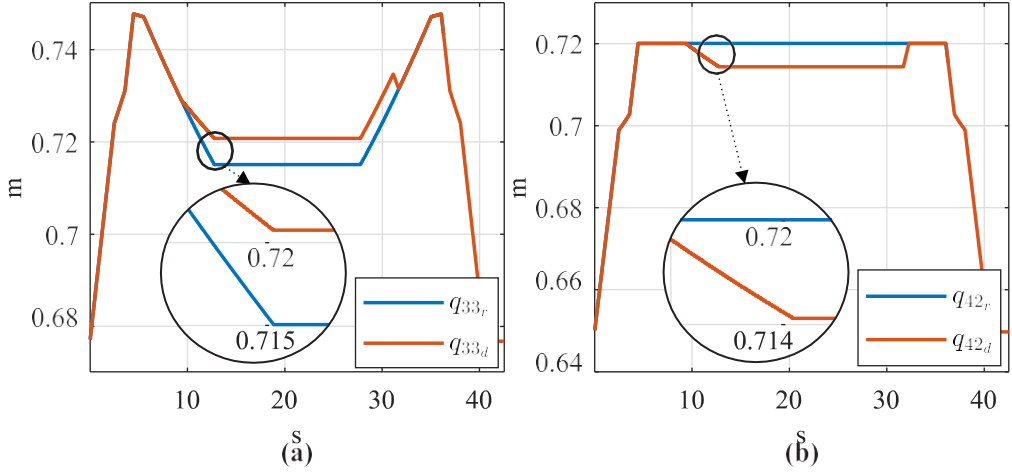

Figure 11: Trajectory planning for the actuator in limb (a) 3 (b) 4.

In this case, the original and the non-singular trajectory are executed in the actual 3UPS+RPU PR while the location and orientation of the mobile platform are measured. The measurement is performed on the point $O_m$ by a 3D tracking system composed of 10 cameras, with an accuracy of 0.5 mm [33]. In Figure 12a, the height measured by the 3D tracking system is plotted for both trajectories. An external perturbation occurs at instant $t = 18s$ where the original trajectory ($z_{m_r}$) yields to the force and moves unexpectedly. In contrast, the trajectory modified by the proposed algorithm ($z_{m_d}$) remains stiff. Moreover, Figure 12b shows the orientation around the tibiofemoral plane for the original ($\theta_r$) and modified trajectory ($\theta_d$).

During the offline trajectory planning, the proposed Type II singularity avoidance algorithm has modified only the trajectory of two actuators with a maximum modification of 6 mm. The offline trajectory planning is implemented in MATLAB on a desktop PC with a processor Core i7 3.7 GHz. In this case, the avoidance algorithm takes on average 1 ms in each iteration. Note that although the proposed algorithm is applied in a spatial case, the execution time is lower than in other algorithms for the planar case, see Table 3. Therefore, these results verify that the Type II singularity avoidance algorithm requires minimum deviation from the original trajectory with low computational cost.

For online trajectory planning, the original trajectory is a hip flexion movement used previously in offline trajectory planning with a different starting location, see Table 6. In this case, the proposed Type II singularity avoidance algorithm analyses each sample of the original trajectory. Then, the non-singular trajectory generated is sent to the PID controller. The PID controller combined with the avoidance algorithm is implemented in ROS2 using the C++ programming language at a rate of 100 Hz. The signal $x_m$ required by the Type II singularity avoidance al-gorithm is provided by a 3D tracking system at a rate of 120 Hz. As in the offline trajectory planning, the singularity avoidance algorithm is set to $t_s$ = 10 *ms* and $v_d$ = 0.01 *m/s*.

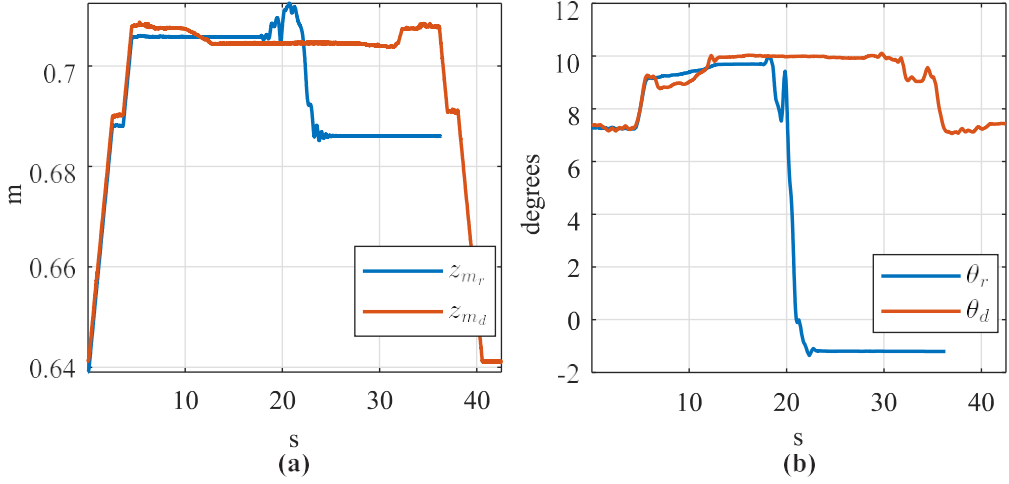

Figure 12: Tracking for PR pose: (a) location on $z_m$ (b) rotation around $\theta$.

Table 6: Description of the trajectory for online planning in the 4-DOF PR.

| Location | $x_m$ (m) | $z_m$ (m) | $\theta$ (degrees) | $\psi$ (degrees) | time (s) |
|---|---|---|---|---|---|
| Start | 0.170 | 0.668 | 12.560 | 8.70 | 0 |
| Singularity | 0.016 | 0.707 | 8.619 | 18.15 | 16.35 |
| End | 0.170 | 0.668 | 12.560 | 8.70 | 47.69 |

Figure 13a presents the original trajectory for the actuator 3 ($q_{33_r}$) and the non-singular trajectory $q_{33_d}$ calculated online. Figure 13b presents the original reference for the actuator 4 ($q_{42_r}$) and the non-singular trajectory $q_{42_d}$. This figure verifies that the proposed algorithm introduces a maximum modification of 7 mm in actuators 3 and 4 during an online Type II singularity avoidance.

Figure 14a presents the height for the original ($z_{m_r}$) and the height calculated online by the proposed algorithm ($z_{m_d}$). Figure 14b shows the orientation around the tibiofemoral plane for the original ($\theta_r$) and the orientation modified by the proposed algorithm ($\theta_d$). Figure 14 shows that the original trajectory on the configuration space is modified maximum 7 mm and 1.5 °. The Type II singularity avoidance algorithm for online trajectory planning is implemented in an industrial PC with a processor Core i7 3.4 GHz. The proposed avoidance algorithm requires on average 3.86 ms at each iteration during the online trajectory planning, and the PID controller is executed in 2 ms. If the controller is executed every 10 ms, the avoidance algorithm employs 38.6 % of the control period and the PID controller 20 %, i.e., 41.4 % of the control period is free. Thus, these results verify that the Type II singularity avoidance algorithm is suitable for online trajectory planning.

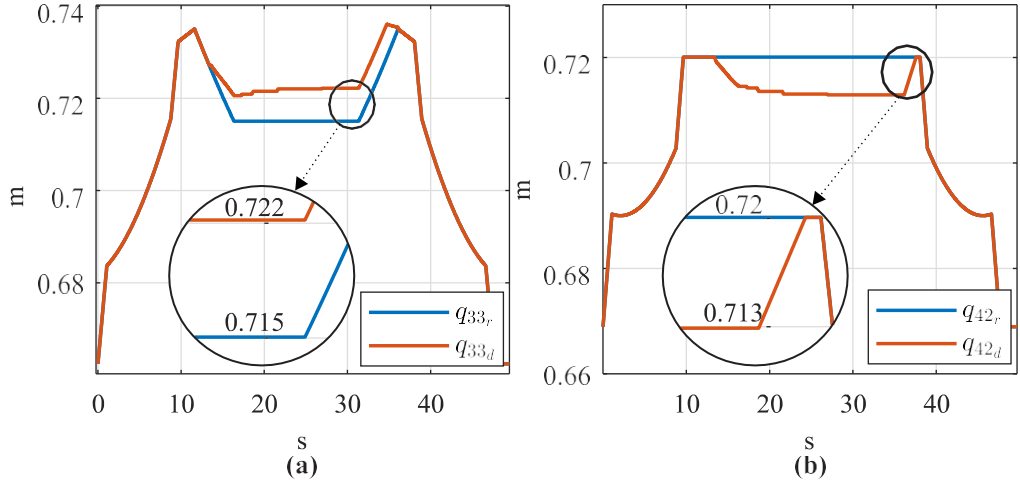

Figure 13: Results of online trajectory planning for the actuator in limb (a) 3 (b) 4.

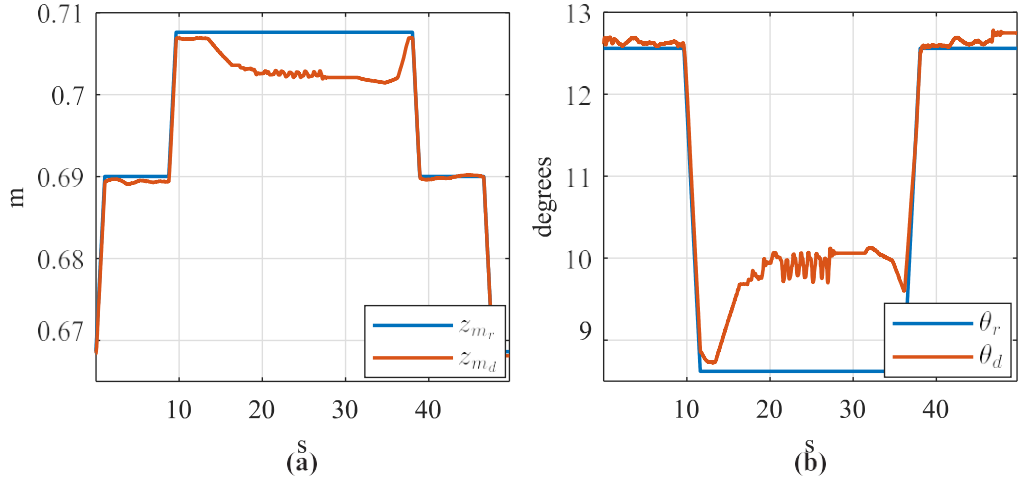

Figure 14: Results of online trajectory planning: (a) location on $z_m$ (b) rotation around $\theta$.

## 6. Conclusions

Based on the angles between the linear part $\mathbf{\Theta}_{i,j}$ and the angular part $\mathbf{\Omega}_{i,j}$ of two OTSs, a Type II singularity avoidance algorithm has been developed for planar and spatial PRs. The singularity avoidance is achieved by modifying the trajectory of the two actuators identified by the minimum angles $\mathbf{\Theta}_{i,j}$ and $\mathbf{\Omega}_{i,j}$ for the planar and spatial case, respectively. The Type II avoidance algorithm has been successfully applied in offline trajectory planning for a symmetrical 5R mechanism (planar case) and a 3U̲PS+RP̲U PR (spatial case). Moreover, the proposed avoidance algorithm has been applied in online trajectory planning for the 3UPS+RPU PR. The 3UPS+RPU PR has been analysed without human interaction to avoid risking the integrity of the patient. Table 7 summarises the results of the Type II avoidance algorithm during offline and online trajectory planning. In Table 7, $t_I$ represents the execution time in each iteration, $\Delta_q$ stands for the maximum modification in the trajectory of actuators, and $\Delta_{\dot{q}}$ is the average deviation in the velocity reference in joint space.

Table 7: Results for the proposed avoidance algorithm for planar and spatial PRs.

| Planning type | PR | $t_I$ (ms) | $\Delta_q$ | $\Delta_{\dot{q}}$ |
|---|---|---|---|---|
| Offline | 5R | 0.5 | 1.2 ° | 0.58 °/s |
| Offline | 3UPS+RPU | 1 | 6 mm | 0.24 mm / s |
| Online | 3UPS+RPU | 3.86 | 7 mm | 0.28 mm / s |

The Type II singularity avoidance algorithm requires 1 ms in each iteration during offline trajectory planning of spatial case, see Table 7. The proposed algorithm was compared with a constrained multi-objective (CMO) algorithm and roadmap (C-space) algorithm for offline trajectory in planar PRs. The proposed avoidance algorithm applied to a 4-DOF PR is faster than CMO and C-space algorithms applied to planar cases. Thus, the low computational cost of the Type II singularity avoidance algorithm is verified. During the online trajectory planning for a 3UPS+RPU PR, the Type II avoidance algorithm requires 3.86 ms for execution. Considering that the controller is executed at every 10ms and the PID law takes 2 ms, there is 4.41 ms free at each iteration. Therefore, the Type II singularity avoidance algorithm is suitable for online applications in the 3UPS+RPU PR.

In offline trajectory planning with a 5R mechanism, the proposed avoidance algorithm introduced a maximum deviation of 2 ° that is imperceptible for the planar PR under study. The velocity trajectory was modified 0.58 °/s, i.e., 2 % of the working velocity. In offline trajectory planning with the 3UPS+RPU PR, the trajectories of the two actuators were modified by a maximum of 6 mm, and the velocity profile changed on average 0.24 mm / s (2.4 % of the working velocity). Considering that the 3UPS+RPU PR is applied for knee rehabilitation, 6 mm is a minimum deviation compared with the range of movement of the human leg. In addition, the proposed algorithm modifies the trajectory of two actuators with a maximum deviation of 7 mm during the online trajectory planning in the 3UPS+RPU PR. Thereby, the proposed algorithm requires a minimum deviation on two actuators to avoid a Type II singularity during offline and online trajectory planning for the 5R and 3UPS+RPU PRs. The minimum deviation on the original trajectory and the low computation cost make the Type II singularity avoidance algorithm suitable for knee rehabilitation assisted by the 3UPS+RPU PR. Note that the proposed avoidance algorithm modifies the trajectory of two actuators because in the proximity to a Type II singularity at least two actuators transmit motion in the same direction. Three actuators contributing in the same direction appear when the Type II singularity is already reached.

The proposed Type II avoidance algorithm requires little tuning for implementation. The proposed avoidance algorithm is tuned by three comprehensible parameters: the sample time ($t_s$), the avoidance velocity ($v_d$) and the limit for the proximity to a singularity ($lim_\alpha$). The $t_s$ is defined by the controller sample time. The $v_d$ is set by the maximum working velocity of the actuators according to the PR application. The $lim_\alpha$ is calculated by trial and error with trajectories with a Type II singularity as the final location. The experimental setting of $lim_\alpha$ allows the incorporation of non-modelled effects such as join clearances and manufacturing errors which increase the accuracy of the avoidance algorithm in real implementations. In addition, the Type II avoidance algorithm only requires the previous reference location during offline trajectory planning. This suppresses the necessity of prior information about the workspace of the PR. The possibility to avoid Type II singularity online reduces the optimisation procedures during the design of PRs and allows to integrate PRs into human interaction tasks with minimum risks.

This research extends the application of the proposed avoidance algorithm for spatial PRs presented by the authors in [34] to planar PRs. In addition, for the first time, the minimum angle $\Theta_{i,j}$ is applied as a proximity detector of Type II singularities for a planar PR during offline trajectory planning. The analysis of the angle between the linear part of two OTSs ($\Theta_{i,j}$) complements the research developed for proximity detection to a Type II singularity based on the angle $\Omega_{i,j}$ [32].

In future work, the proposed algorithm is going to be combined with a force controller to perform active and passive exercises for knee rehabilitation using the 3U$\underline{P}$S+R$\underline{P}$U PR. The force controller requires an online Type II singularity avoidance algorithm because the user could drive the PR to a singular configuration by accident. In this case, the actual location of the PR should be measured by a redundant 3D tracking system or by solving the Forward kinematics based on embedded encoders. The implementation of the proposed avoidance algorithm was limited to offline trajectory planning for the 5R mechanism and the 3U$\underline{P}$S+R$\underline{P}$U PR. Moreover, the implementation of the index $\Theta_{i,j}$ in online trajectory planning has not been analysed. Therefore, the Type II singularity avoidance algorithm should be tested in PRs with different architectures to generalise the features discussed in this research.

**Acknowledgements**

This research was partially funded by Fondo Europeo de Desarrollo Regional (PID2021-125694OB-I00), cofounded by Vicerrectorado de Investigación de la Universitat Politècnica de València (PAID-11-21) and by Programa de Ayudas de Investigación y Desarrollo de la Universitat Politècnica de València (PAID-01-19).

Funding for open access charge: CRUE-Universitat Politècnica de València.

Moreover, the authors would like to thank the help of all anonymous reviewers, which have improved the paper's readability.